\documentclass{article}
\usepackage[utf8]{inputenc}
\usepackage{color}
\usepackage{graphicx}
\usepackage{dingbat}
\usepackage{pdflscape}
\usepackage{longtable}
\usepackage{authblk}
\usepackage{multirow}
\usepackage{multicol}
\usepackage{tabularx}
\usepackage{arydshln}
\usepackage{todonotes}
\usepackage{caption}
\usepackage{subcaption}
\usepackage{soul}
\usepackage{geometry}
\usepackage{setspace}
\usepackage{sectsty}
\usepackage{titlecaps}
\sectionfont{\MakeUppercase}
\usepackage{hyperref}
\hypersetup{colorlinks=true, linkcolor=blue, citecolor=blue, urlcolor=black, plainpages=false, pdfpagelabels, hypertexnames=false, unicode}

\usepackage[symbol]{footmisc}
\usepackage{wrapfig}

\newcommand{\beginsupplement}{%
        \setcounter{table}{0}
        \renewcommand{\thetable}{S\arabic{table}}%
        \setcounter{figure}{0}
        \renewcommand{\thefigure}{S\arabic{figure}}%
     }

\title{\textbf{Extracting Social Support and Social Isolation Information from Clinical Psychiatry Notes: Comparing a Rule-based NLP System and a Large Language Model}}

\author[1\thanks{co-first authors}]{Braja Gopal Patra}
\author[2*]{Lauren A. Lepow}
\author[1]{Praneet Kasi Reddy Jagadeesh Kumar} 
\author[1]{Veer Vekaria}
\author[1]{Mohit Manoj Sharma}
\author[1]{Prakash Adekkanattu}
\author[2]{Brian Fennessy}
\author[2]{Gavin Hynes}
\author[2]{Isotta Landi}
\author[3]{Jorge A. Sanchez-Ruiz}
\author[3]{Euijung Ryu}
\author[3]{Joanna M. Biernacka}

\author[2]{Girish N. Nadkarni}
\author[4,5]{Ardesheer Talati}
\author[4,5]{Myrna Weissman}
\author[4,5,6]{Mark Olfson}
\author[6]{J. John Mann}
\author[2]{Alexander W. Charney}
\author[1]{Jyotishman Pathak}

\affil[1]{Weill Cornell Medicine, New York, NY, USA}
\affil[2]{Icahn School of Medicine at Mount Sinai, New York, NY, USA}
\affil[3]{Mayo Clinic, Rochester, MN, USA}
\affil[4]{Columbia University Vagelos College of Physicians and Surgeons, New York, NY, USA}
\affil[5]{New York State Psychiatric Institute, New York, NY, USA}
\affil[6]{Columbia University Irving Medical Center, New York, NY, USA}

\begin{document}
\doublespacing
\maketitle
\newpage

\begin{abstract}
\noindent{\bf Background:} Social support (SS) and social isolation (SI) are social determinants of health (SDOH) associated with psychiatric outcomes. In electronic health records (EHRs), individual-level SS/SI is typically documented as narrative clinical notes rather than structured coded data. Natural language processing (NLP) algorithms can automate the otherwise labor-intensive process of data extraction.


\noindent{\bf Data and Methods:} Psychiatric encounter notes from Mount Sinai Health System (MSHS, n=300) and Weill Cornell Medicine (WCM, n=225) were annotated and established a gold standard corpus. A rule-based system (RBS) involving lexicons and a large language model (LLM) using FLAN-T5-XL were developed to identify mentions of SS and SI and their subcategories (e.g., social network, instrumental support, and loneliness).  


\noindent{\bf Results:} For extracting SS/SI, the RBS obtained higher macro-averaged f-scores than the LLM at both MSHS (0.89 vs. 0.65) and WCM (0.85 vs. 0.82). For extracting subcategories, the RBS also outperformed the LLM at both MSHS (0.90 vs. 0.62) and WCM (0.82 vs. 0.81).

\noindent{\bf Discussion and Conclusion:} Unexpectedly, the RBS outperformed the LLMs across all metrics. Intensive review demonstrates that this finding is due to the divergent approach taken by the RBS and LLM. The RBS were designed and refined to follow the same specific rules as the gold standard annotations. Conversely, the LLM were more inclusive with categorization and conformed to common English-language understanding. Both approaches offer advantages and are made available open-source for future testing.

\end{abstract}

\section{Introduction}

Social determinants of health (SDOH) are the non-medical conditions that shape daily life and affect health outcomes~\cite{who}. Leveraging SDOH in clinical decision-making may personalize treatment planning and improve patient outcomes~\cite{patra2021extracting}. Social support (SS) and social isolation (SI) are two key components of SDOH that significantly impact physical and mental well-being.

SI is associated with higher health care expenditure~\cite{polikandrioti2022perceived}, morbidity~\cite{zhu2021living,lin2022impact,olano2022causes}, and mortality~\cite{cene2022effects,naito2023mortality} and may be as harmful as smoking fifteen cigarettes a day~\cite{general2023our}. Specific health risks liked to SI include poor physical and mental well-being~\cite{cacioppo2014social}, metabolic diseases, infectious diseases, dementia~\cite{general2023our}, suicidal thoughts~\cite{kim2021social}, anxiety~\cite{wickramaratne2022social}, and depression~\cite{kim2021social,czaja2021social}. We previously conducted a scoping review to evaluate the relationship between social connectedness and the risk of depression or anxiety, and observed that loneliness was significantly associated with higher risks of major depressive disorder, depressive symptom severity, and generalized anxiety disorder~\cite{wickramaratne2022social}. SI comprises several interrelated psychosocial constructs including a lack of a social network, poor emotional support, and feelings of loneliness~\cite{lamblin2017social}. The Surgeon General’s 2023 advisory, ``Our Epidemic of Loneliness and Isolation,'' recommends identification of patients with SI at the health care system-level to track community prevalence. Research may then study causal mechanisms, patterns across demographics, and preventive approaches~\cite{general2023our}.

In contrast, SS and related constructs including emotional support, instrumental support, and social network are associated with improved health outcomes~\cite{purcell2021effectiveness,lin2022impact,freak2021social}. Social connections facilitate health-related behaviors, including adherence to medication and treatment~\cite{dimatteo2004social,dimatteo2004variations}. Moreover, social relationships are indirectly associated with several aspects relevant to health, such as blood pressure, immune function, and inflammation~\cite{robles2003physiology,uchino2006social,hostinar2014psychobiological,holt2017potential}. In regards to mental health, SS, when measured across a range of settings and populations and using a variety of measures, may be a protective factor for depressive symptoms and disorders~\cite{wickramaratne2022social}. 

Existing studies on SS and SI are largely based on questionnaire or survey data from small samples or specific populations (e.g., the elderly~\cite{czaja2021social}, adolescents during the pandemic~\cite{trucco2023social,thakur2023adolescent}, and pregnant/postpartum women~\cite{wickramaratne2022social}). Research on identifying SS and SI from real world, large-scale, routinely collected electronic health records (EHRs) is lacking, likely because SDOH information, including SS and SI, are rarely encoded in EHRs as structured data elements~\cite{patra2021extracting}. International Classification of Diseases (ICD)-9 V-codes and ICD-10 Z-codes have been expanded to include SI, however, studies note poor adoption rates among clinicians and health systems~\cite{truong2020utilization}. Instead, these concepts are often captured in EHRs as part of narrative text during a clinical encounter, yet manual abstraction of such data is time-consuming and labor intensive~\cite{zhu2019automatically,patra2021extracting}. 

Natural language processing (NLP) automates extraction of information from unstructured data and has been implemented in previous literature for identifying different SDOH constructs, including alcohol use, substance use, and homelessness~\cite{patra2021extracting}. However, the highly varied language used by clinicians, domain/site-specific knowledge, and lack of annotated data present challenges in extracting SDOH from clinical notes~\cite{patra2021extracting}. There are currently three main approaches to extract SS and SI in clinical text, each with strengths and limitations. Note that existing NLP work on SDOH have yet to be optimized for capturing SI/SS.

The first approach involves creating dictionaries (``lexicons'') and a set of rules with which to search the text for matches. Lexicons may be either derived from standardized medical ontologies or developed specifically for the task by domain experts. Software may be used to implement the rules, including recognition of negative terms or contexts in which the lexicon match is a false-positive (e.g., if the documented SDOH is not describing the patient, but rather the patient's sibling). The benefit of this method is that the parameters are highly controlled; there is no ``black box'' (an inability to see how the model makes decisions) since the pipeline creator is naming exactly what is, or is not, included. However, it is exceedingly difficult to list every term in the lexicon and create a rule for every \textit{context} in which the term might occur. Previous work using this approach includes studies detecting SI from clinical notes~\cite{zhu2019automatically,kharrazi2018value} in specific patient populations. ClinicalRegex and Linguamatics I2E, two rule-based/lexicon software, were used to extract SI~\cite{greenwald2017novel} and SS mentions, again in specific patient populations~\cite{bhatt2023use}. Other studies [e.g., Navathe et al.~\cite{navathe2018hospital}] combine ICD codes with lexicon terms to detect SS/SI. Since the aim of these studies is to identify SS/SI for a clinical purpose, the focus is not on the rigor of algorithm development; therefore, they take more blunt approaches, such as (a) not differentiating between types of SS/SI or considering nuances, (b) using a relatively small sample of manually validated notes, (c) using a single site, and (d) not typically making their pipelines publicly available.

The second NLP approach involves traditional training or adapting various machine learning models (pre-packaged topic modeling, deep learning, and language models). EHR-based research has used models trained on a clinical corpus and thus are ideally suited to understand clinical language. However, to perform a task such as identifying specialized concepts like SS/SI, these models still require extensive manually labeled training data to fine-tune the model, which is labor-intensive and generates results that underperform the lexicon-based approach~\cite{wang2018disease}. 

Finally, an emerging approach is to use large language models (LLMs) which have been trained on massive amounts of data, and use transfer learning to perform downstream tasks with little need for fine-tuning or manual labels. The advent of LLMs is a major milestone in the field of NLP and has been used for several tasks in health informatics including SDOH extraction from clinical notes~\cite{guevara2023large,lybarger2023advancements}. Preliminary work has used LLMs with minimal fine-tuning to extract SDOH, but the performance of identifying SS/SI has yet to be optimized for clinical or research applications~\cite{guevara2023large}.

In summary, each of these approaches for extracting SDOH from clinical text have strengths and limitations. The rule-based system (RBS) requires domain experts and significant time to develop lexicons and rules, which results in highly predictable outputs. In contrast, machine learning and deep learning-based systems rely heavily on a large, annotated corpus for training. Lastly, LLMs need less data for fine-tuning compared to deep learning algorithms, but are often considered black-box models, making their decision-making processes less transparent.

This work aims to build on these previous systems by breaking down SS/SI into their fine-grained categories including presence/absence of \textit{social network}, \textit{instrumental support}, \textit{emotional support}, and \textit{loneliness}. This separation is important, as the literature has shown that they are separate concepts~\cite{barrera1986distinctions}, not interchangeable, with distinct effects on health outcomes~\cite{wickramaratne2022social}. A general label not only diminishes the signal of detectable associations between subcategories, but also limits the eventual interventions that might come from findings. A distinction is frequently drawn between subjective and objective social support, and they do not necessarily improve together~\cite{ge2017social}. For example, loneliness is frequently found to be associated with depressive symptoms, but increasing a person’s social activity is not necessarily the way to alleviate loneliness, and other interventions might be more indicated for the individual experiencing loneliness~\cite{masi2011meta}. This study aims to fill a gap in the literature by not only focusing on SS/SI extraction in clinical narratives, but also distinguishing between classes. Here, we describe the development of a rule book for manual annotations as well as rigorous development of a rule-based system (RBS) and an LLM-based system (LLM).

Additionally, the variability in clinical documentation, both within and across hospital systems, presents a challenge to the portability of NLP systems~\cite{patra2021extracting}, and previously published lexicons and pipelines were created for single EHR datasets. An additional aim of this work is to create NLP pipelines that are portable across sites, here, two large academic medical centers in New York City: Mount Sinai Health System [MSHS] and Weill Cornell Medicine [WCM]. By making benchmarked NLP pipelines open-source, we aim to enable other healthcare systems to adopt, validate, improve, and deploy the developed SI/SS extraction tool for contextualizing both psychiatric research and clinical practices.






%





\section{Data}

\subsection{Data Sources}
\label{datasource}

We used EHR data from two sites (MSHS and WCM) to develop the SS/SI NLP pipeline. First, we used 286,692 notes from 33,800 patients derived from the MSHS data warehouse. Notes are from psychiatric inpatient, psychiatric emergency department (ED), and psychiatry consult-liaison encounters between 2011 and 2020. Advantages of the corpus include relative consistency in documentation style and comprehensive biopsychosocial patient evaluations. Of note, the corpus also contains the occasional presence of clinical note templates (found in 14\% of the notes) such as:`lacks social support: yes/no/unknown.' For this study, templates were removed while annotating and evaluating the NLP system; however, in the future, we will compare clinician responses in these templates to the gold-standard annotations.

Second, we used 48.98 million clinical notes from 558,133 patients derived from the WCM enterprise data warehouse. These notes from 2010 to 2023 comprise patients with any mental health diagnosis or antidepressant prescription from outpatient, inpatient, or ED visits. To align WCM data with the MSHS data, NLP development was restricted to psychiatric encounter notes only. The study was approved by the Institutional Review Boards (IRBs) at MSHS and WCM.

\subsection{SS and SI Categories}

\begin{figure}[!htbp]
    \centering
    \includegraphics[width = 12cm]{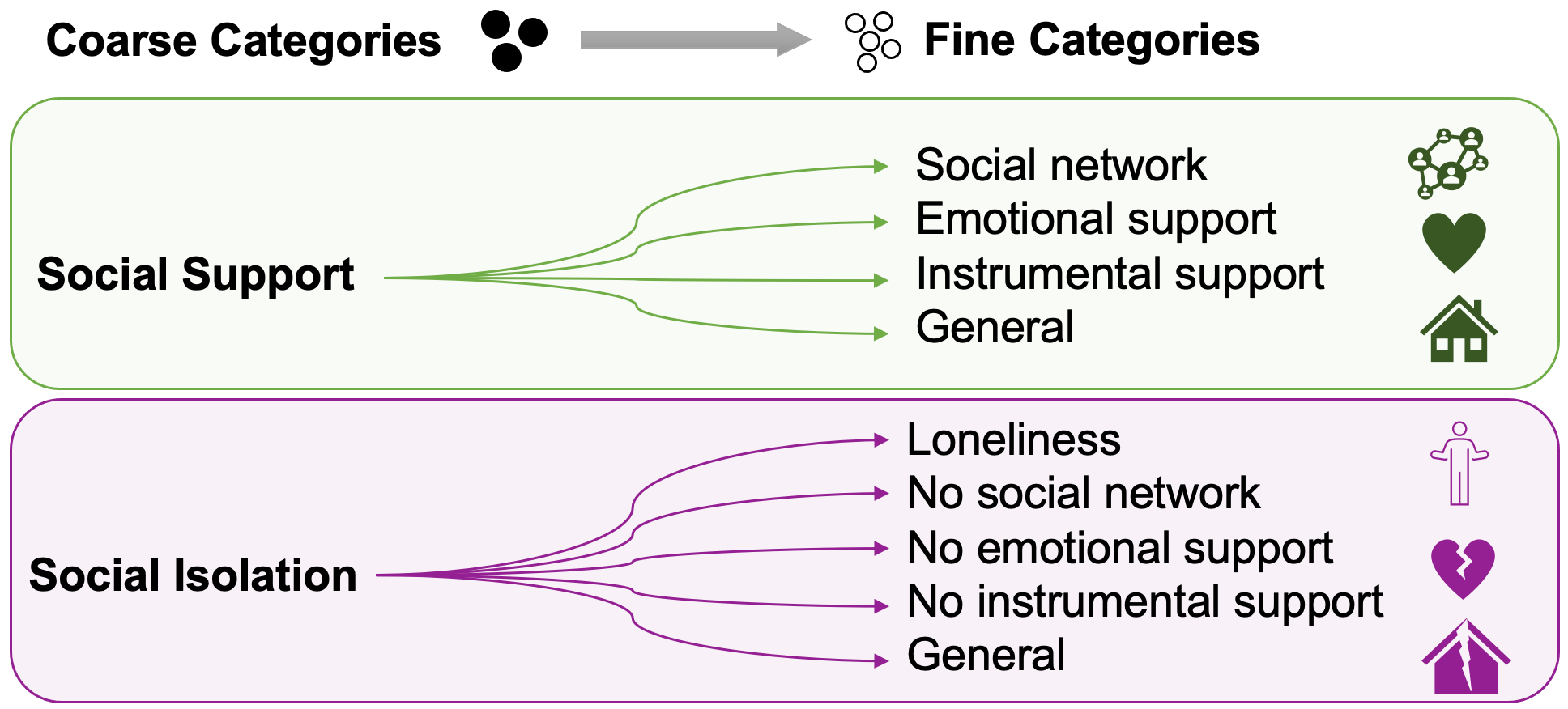}
    \caption{Annotation categories for social support (SS) and social isolation (SI)}
    \label{schema}
\end{figure}

In addition to the two coarse-grained categories (SS and SI), we sought to further classify these concepts into distinct fine-grained categories that uniquely impact health and wellbeing~\cite{heaney2008social}. The fine-grained categories are based on the seminal work of House et al.~\cite{house1987social} and updated to include what has been researched in healthcare (e.g., social networks)~\cite{wickramaratne2022social}. Our workgroup of clinical psychiatrists and psychologists, psychiatric epidemiologists, sociologists, and biomedical informaticians finalized the categories. There is an inherent degree of overlap and subjectivity between the fine-grained categories, e.g., `lives with family' could conceivably be characterized as \textit{instrumental support}, \textit{emotional support}, \textit{social network}, or perhaps a \textit{general} category of SS. Therefore, to distinguish between categories we created a clear annotation rule book~\cite{house1987social} to ensure consistency, transparency, and reproducibility. The final fine-grained SS subcategories included \textit{social network} (`\textit{goes to church}'), \textit{emotional support} (`\textit{can talk about her problems}'), \textit{instrumental support} (`\textit{home health aide}'), and a \textit{general} subcategory\footnote{subcategory and fine-grained category are used interchangeably} (`\textit{patient has good social supports}'), which is assigned when there are insufficient details to ascertain a category. The final fine-grained SI subcategories included \textit{loneliness} (`\textit{feelings of loneliness}'), \textit{no social network} (`\textit{socially isolated}'), \textit{no emotional support} (`\textit{no one to confide in}'), \textit{no instrumental support} (`\textit{homeless}'), and a \textit{general} subcategory (`\textit{no social support}'). The schema for SS and SI are presented in Figure~\ref{schema}. 



\section{Methods}

\subsection{Lexicon Creation and Expansion}
\label{lex-creation}

The computational approaches to any NLP tasks require annotated lexicons and gold standard data~\cite{patra2021extracting}. We collected lexicons for fine-grained categories of SS and SI using an iterative method that included manual chart reviews and semi-automatic methods.

\subsubsection{Manual Chart Review}

Zhu et al.~\cite{zhu2019automatically} developed a lexicon for identifying SI from clinical notes of patients with prostate cancer in the context of recovery support. Initially, this lexicon, which included 24 terms, was selected; however, it yielded relatively fewer clinical notes at MSHS and WCM compared to the published report. A list of terms for each category was created and extensively reviewed by  the study team which included clinical psychiatrists and psychologists. We manually reviewed 50 notes at each site to find SS and SI keywords to enrich the existing lexicons.

\subsubsection{Semi-automatic Method}

The lexicons from manual chart review as above were enhanced using word embeddings. First, the manually generated lexicons were vectorized using \texttt{word2vec}~\cite{mikolov2013distributed} and Equation~\ref{eq1}.

\begin{equation}
\label{eq1}
    \vec{lex} = \frac{1}{N} \sum \vec{w}
\end{equation}

Here, $\vec{lex}$ and $\vec{w}$ refer to the lexicon vector and word vector from a lexicon, respectively, and $N$ is the number of words in a lexicon. Then, the top 10 similar keywords were identified for the vector using word2vec's \textit{most\_similar} function. The word2vec model was trained on one million randomly-selected clinical notes separate from the WCM EHR data using the \textit{gensim} package\footnote{https://radimrehurek.com/gensim/}. The new set of keywords was manually reviewed and selected for the next round of vectorization and similarity matching. This process was repeated until the workgroup reached a consensus on the quality of the final set of lexicons.

\subsection{Annotations}
\label{annotation}

To create gold standard data for developing the NLP pipelines, we selected 300 notes from 300 unique patients at MSHS and 225 notes from 221 unique patients at WCM for fine- and coarse-grained manual annotation. Notes were chosen from unique patients to maximize the contextual diversity of SS/SI terms (different note-writers, different time periods, and avoiding redundancy caused by copy-forward practices within a single patient’s EHR). To optimize the gold standard annotation set of notes, those selected for review were enriched for mentions of SS and SI: 75 notes were selected that had at least one occurrence of an SI lexicon term, another 75 notes for SS, and finally 75 notes were randomly selected from the reminder of underlying corpus. At MSHS, 75 additional notes were selected that contained a clinical note template to further enrich the annotation corpus for notes in which a clinician was prompted (by the template) to assess SS/SI.


The Brat Rapid Annotation Tool (BRAT)~\cite{brattool} was used to annotate the notes manually with the same annotation configuration schema across sites. The annotation guideline\footnote{rule book and annotation guideline are used interchangeably} and lexicons are provided in Supplementary Tables~\ref{lexicons}. Initially, the annotations were performed at the entity level (every instance of a lexicon term in the note text) using BRAT. For evaluation, the entity-level annotations were converted to ``document'' (note) level. For example, if there was a single entity mentioning \textit{loneliness} and two mentions of \textit{instrumental support} in a given note, the \textit{loneliness} and \textit{instrumental support} subcategories were assigned to that note. Finally, the coarse-grained categories were assigned to each document using rules. SS was assigned to a document if there were one or more mentions of any SS subcategories and similarly, SI was labeled if there were one or more mentions of any SI subcategories. The above note would be annotated with both SI (for \textit{loneliness}) and SS (for \textit{instrumental support}).

The notes were meticulously reviewed by two annotators and disagreements were resolved by a third adjudicator to create the final gold-standard corpus. For coarse-grained annotation, the inter-annotator agreement (IAA) Cohen's Kappa scores were 0.92 [MSHS] and 0.86 [WCM]; for fine-grained, 0.77 [MSHS] and 0.81 [WCM]. The counts of fine- and coarse-grained categories found in the gold-standard data are provided in Supplemental Table~\ref{tab:iaa:dataset}.

The rule book was used to train the annotators and were continually updated during the adjudication process. Often, disagreeing annotations could both be seen as correct given the inherent subjectivity of the classification process; however, new rules were created to arrive at one consistent label for edge cases. Sometimes, rules were created for more practical reasons, for example, mentions of `\textit{psychotherapy}' were excluded from \textit{emotional support} because otherwise almost every note in the MSHS psychiatric corpus would be flagged. Of note, mentions were only labelled when SS/SI was explicit and not implied. For example, a mention of `\textit{boyfriend}' or `\textit{living alone}' without further context would not count. The \textit{general} subcategory became a ``catch-all'' for mentions that clearly involved support or isolation, but a single fine-grained category could not be discerned. For example, `\textit{staying on his best friend's couch}' could be seen as \textit{instrumental support} (providing shelter), \textit{social network} (having friends), or \textit{emotional support} (best friend implies a level of closeness). At both institutions, the IAA was reflective of the subjective, overlapping nature of the fine-grained subcategories. Another reason for disagreements between annotators was the site-specific familiarity required to recognize acronyms and social services, e.g., `\textit{HASA stands for the HIV/AIDS Services Administration}.' 

\subsection{System Description}
\begin{figure}[!htbp]
    \centering
    \includegraphics[width = 14cm]{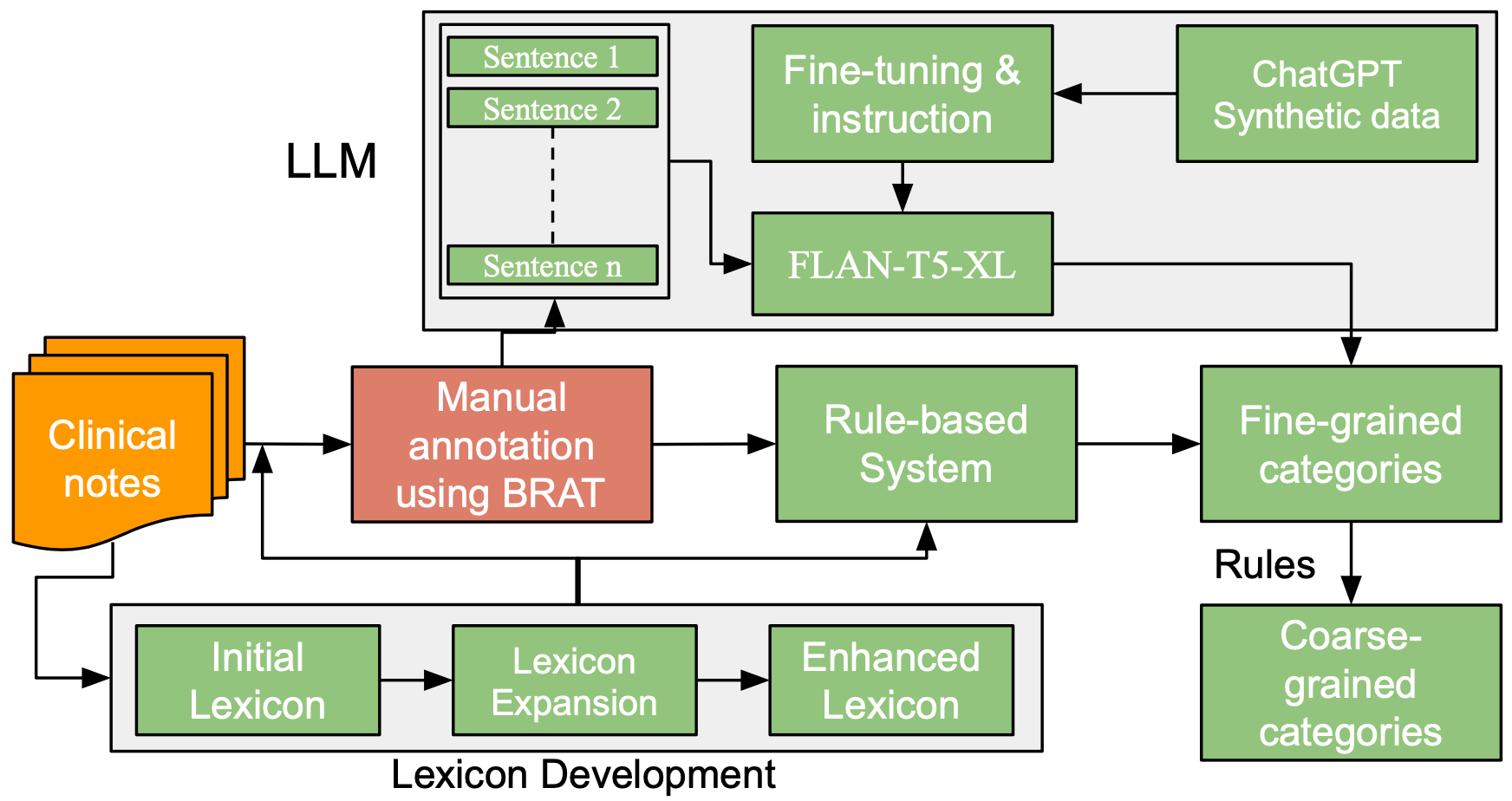}
    \caption{Architecture of the rule- and large language model (LLM)-based NLP systems for identifying fine- and coarse-grained categories. For the LLM input, a single clinical note was sliced into multiple sentences due to the restriction of 512 tokens. The sentence-level fine-grained categories were combined to provide document-level fine-grained categories. Finally, the rules in Section~\ref{annotation} were used to identify the coarse-grained categories from fine-grained categories.}
    \label{fig:arch_nlp}
\end{figure}

We developed rule- and LLM-based systems to identify mentions of fine-grained categories in clinical notes. Rules were then used to translate entity-level to note-level classifications and fine-grained to coarse-grained labels as mentioned in Section~\ref{annotation}. An architecture of the NLP systems is provided in Figure~\ref{fig:arch_nlp}. 

\subsubsection{Rule-based System}

As aforementioned, a major advantage of the RBS is full transparency in how classification decisions are made. 
We implemented the system using the open source \textit{spacy Matcher}\footnote{https://spacy.io/api/matcher}\cite{spacy}. Additionally, we compiled a list of exclusion keywords (see Supplementary Table~\ref{exclusion_terms}) to refine the rules, ensuring relevant identification.


 


\subsubsection{Supervised Models} 

Expanding on the published literature, we attempted first we began to implement Support Vector Machines (SVMs) and Bidirectional Encoder Representations from Transformers (BERT)-based models at WCM to identify fine-grained categories. However, these models were inappropriate due few SS/SI mentions in the corpus (see Supplementary Material and Table~\ref{tab:mldl_results}). 

\subsubsection{Large Language Models (LLMs)} 

We developed a semi-automated method to identify SS and SI using an open-source advanced fine-tuned LLM called ``Fine-tuned LAnguage Net-T5 (FLAN-T5)"~\cite{raffel2020exploring,chung2022scaling}. We used FLAN-T5 in a ``question-answering" fashion to extract sentences from clinical texts with mentions of SS and SI subcategories. A separate fine-tuned model was created for each of the fine-grained categories.

\textbf{Model Selection:} T5 has been used for other classification tasks in clinical notes, and the FLAN (Fine-tuned Language Net) version of T5, which employs chain-of-thought (COT) prompting, does not require labeled training data~\cite{chung2022scaling}. Five variants of FLAN-T5 are available based on the number of model parameters\footnote{https://huggingface.co/docs/transformers/model\_doc/flan-t5}. 
Guevara et al.~\cite{guevara2023large} observed that FLAN-T5-XL performed better than the smaller models (FLAN-T5-L, FLAN-T5-base, and FLAN-T5-small) with no significant improvement with the larger FLAN-T5-XXL. Thus, we selected FLAN-T5-XL for our experimentation.

\textbf{Zero-shot:} Given that LLMs follow instructions and are trained on massive amounts of data, they do not necessarily require labeled training data. This ``zero-shot" approach was performed by providing model \textit{instruction}, \textit{context}, a \textit{question}, and possible \textit{choice} (`yes,' `no,' or `not relevant'). An example is provided in Table~\ref{tab:example}. The option `no' was selected for contexts that were negated, and `not relevant' was chosen for those that did not pertain to the subcategory or the question.


\textbf{Fine-tuning:} Since FLAN-T5-XL (zero-shot) with instruction had poor F-scores (see Supplementary Table~\ref{llm-result}), the models were improved by fine-tuning them with synthetic examples that could help the model learn about the specific SS or SI subcategories. For each fine-grained category, about 50 (yes), 50 (no), and 50 (not relevant) examples were created. The synthetic examples themselves became a validation set to fine-tune the parameters. ChatGPT (with GPT 4.0)\footnote{https://openai.com/blog/chatgpt} was used to help craft context examples, but ultimately after several iterations in the validation set, they were refined by the domain experts so that each example was specifically instructive about the inclusions and exclusions of the category. Examples of prompts for \textit{loneliness} are provided in Table~\ref{tab:example}. All fine-tuning examples and questions for each subcategory are provided in Supplementary Material and Table~\ref{questions}. Furthermore, giving the LLMs specific stepwise instructions to follow (``instruction tuining'') has been shown to improve performance by reducing hallucinations~\cite{ouyang2022training,zhou2023context}. Therefore, we added an instruction as a part of the prompt.   
 
\textbf{Parameters:} Previously, the parameter-efficient Low-Rank Adaptation (LoRA) fine-tuning method was used with FLAN-T5 models to identify SDOH categories~\cite{guevara2023large}. However, the newer Infused Adapter by Inhibiting and Amplifying Inner Activations (IA\textsuperscript{3}) was selected for its better performance~\cite{liu2022few}. We fine-tuned the data on 15-20 epochs. Fine-tuning parameters can be viewed in our publicly available code\footnote{https://github.com/CornellMHILab/Social\_Support\_Social\_Isolation\_Extraction}.

\begin{table}[!htbp]
    \centering 
    \caption{Example of instruction, question, context, choices, and answer for \textit{loneliness} subcategory model.}
    \begin{tabular}{|p{2cm}|p{11cm}|}
        \hline
        \textbf{Instruction} & Read what the Clinician wrote about the patient in the \textit{Context} and answer the \textit{Question} by choosing from the provided \textit{Choices}.\\ \hdashline
        \textbf{Context} & The Clinician wrote: ``\textit{Pt continues to express feelings of loneliness.}'' \\ \hdashline
        \textbf{Question} & In the Clinician's opinion, ``does or did the patient experience feelings of loneliness?"\\ \hdashline
        \textbf{Choices} & yes; no; not relevant \\ \hline
        
        \textbf{Context} & The Clinician wrote: ``\textit{He denies suffering from loneliness.}'' \\
        \textbf{Answer} & no. \\\hline
        \textbf{Context} & The Clinician wrote: ``\textit{Pt is currently homeless.}'' \\
        \textbf{Answer} & not relevant\\\hline
    \end{tabular}
    \label{tab:example}
\end{table}


\subsubsection{Evaluation} 
All evaluations were performed at the note level for both the fine- and coarse-grained categories. To validate the NLP systems, precision, recall, and f-score were all macro-averaged to give equal weight to the number of instances. Instances of \textit{emotional support} and \textit{no emotional support} subcategories were rare in the underlying notes (see Supplementary Table~\ref{tab:iaa:dataset} for full counts) and therefore the accuracy could not be assessed. 

\section{Results}

\subsection{Demographics}
The demographic characteristics of patients within the annotated cohort are detailed in Table~\ref{demographics}. Notably, the patient composition at MSHS was younger and more diverse as compared to patients at WCM.

\begin{table}[!htbp]
    \caption{Patient demographics from annotated data [\# of patients (\%)]. AA: African American}
    \label{demographics}
    \centering
    \begin{tabular}{c|c|c}
        \hline
        \textbf{Characteristics} & \textbf{MSHS} & \textbf{WCM} \\
        \hline
        \multicolumn{3}{l}{\textbf{Age}}\\\hline
        \textbf{$<$18} & 34 (9.04) & 20 (9.05) \\
        \textbf{18-39} & 145 (38.56) & 39 (17.65)\\
        \textbf{40-59} & 120 (31.91) & 82 (37.10) \\
        \textbf{$\geq$60} & 77 (20.48) & 80 (36.20)\\ \hline
        \multicolumn{3}{l}{\textbf{Sex}}\\\hline
        \textbf{Female} & 194 (51.60) & 114 (51.58) \\ 
        \textbf{Male} &182 (48.40)&  107 (48.42)  \\\hline
        \multicolumn{3}{l}{\textbf{Race}}\\ \hline
        \textbf{White} & 86 (28.67) & 88 (39.82) \\ 
        \textbf{Black or AA} & 112 (37.33) & 36 (16.29)  \\ 
        \textbf{Asian} & 7 (2.33) & 5 (2.26)\\
        \textbf{Other} & 95 (31.67)& 50 (22.62) \\
        \textbf{Unknown} & - & 42 (19.00)\\\hline
        \multicolumn{3}{l}{\textbf{Ethnicity}}\\ \hline
        \textbf{Hispanic} & 69 (23.00) & 35 (15.85) \\ 
        \textbf{Non-Hispanic} & 190 (63.33) & 186 (84.16)\\ 
        \textbf{Unknown} & 41 (13.67) & -\\ \hline
    \end{tabular}
\end{table}

\begin{table}[!htbp]
\centering
\caption{Macro-averaged Precision (P), Recall (R), and F-scores (F) of different NLP pipelines for fine and coarse-grained category classification. Here, we used fine-tuning and instruction for FLAN-T5-XL model. The highest scores for individual categories are underlined.}
\label{main-result}
\begin{tabular}{cl|ccc|ccc}
\hline
\multicolumn{2}{c}{\textbf{Categories}} & \multicolumn{3}{|c|}{\textbf{Rule}} & \multicolumn{3}{c}{\textbf{FLAN-T5-XL}} \\ \cline{3-8}
& & \textbf{P} & \textbf{R} & \textbf{F} & \textbf{P} & \textbf{R} & \textbf{F} \\\hline

 & \multicolumn{7}{c}{\textbf{MSHS}} \\\hline

\multirow{5}{*}{\textbf{SS}} & Soc. Network & 0.84 & 0.83 & \underline{0.84} & 0.66 & 0.70 & 0.60\\
 & Emotional  & 0.49 & 0.49 & \underline{0.49}  & -- & -- & -- \\
 & Instrumental &  0.94 & 0.91 & \underline{0.92} & 0.60 & 0.64 & 0.43 \\
 & General & 0.86 & 0.82 & \underline{0.83} & 0.65 & 0.69 & 0.57 \\
 \hdashline
\multicolumn{1}{c}{\multirow{6}{*}{\textbf{SI}}} & Loneliness & 0.96 & 0.97 & \underline{0.97} & 0.65 & 0.89 & 0.69\\
 & no Soc. Network & 0.92 & 0.94 & \underline{0.93}  & 0.63 & 0.83 & 0.65\\
 & no Emotional & 1.00 & 1.00 & \underline{1.00} & -- & -- & -- \\
& no Instrumental & 0.97 & 0.91 & \underline{0.94} & 0.63 & 0.76 & 0.55 \\
& General & 0.93 & 0.92 & \underline{0.93} & 0.77 & 0.83 & 0.79\\
 & Average & 0.89 & 0.91 & \underline{0.90} & 0.65 & 0.76 & 0.62\\\hline

 & SS & 0.85 & 0.85 & \underline{0.84} & 0.75 & 0.61 & 0.55  \\
 & SI & 0.95 & 0.95 & \underline{0.95} & 0.80 & 0.76 & 0.72 \\
 & Average & 0.90 & 0.89 & \underline{0.89} & 0.78 & 0.69 & 0.65\\\hline
 
 & \multicolumn{7}{c}{\textbf{WCM}} \\\hline
 
\multirow{5}{*}{\textbf{SS}} & Soc. Network & 0.83 & 0.82 & \underline{0.82} & 0.81 & 0.83 & \underline{0.82} \\
 & Emotional & 0.62 & 0.80 & \underline{0.60} & -- & -- & -- \\
 & Instrumental & 0.81 & 0.75 & 0.75 & 0.82 & 0.81 & \underline{0.80}\\
 & General & 0.93 & 0.88 & \underline{0.90} & 0.77 & 0.84 & 0.77\\
 \hdashline
\multirow{6}{*}{\textbf{SI}} & Loneliness & 0.99 & 0.95 & \underline{0.97} & 0.84 & 0.93 & 0.88\\
 & no Soc. Network & 0.93 & 0.69 & 0.74 & 0.81 & 0.88 & \underline{0.75}\\
 & no Emotional & 0.98 & 0.62 & \underline{0.69} & -- & -- & -- \\
 & no Instrumental & 0.93 & 0.83 & \underline{0.87} & 0.76 & 0.81 & 0.77\\
 & General & 0.94 & 0.89 & \underline{0.91} & 0.81 & 0.80 & 0.80\\
 & Average & 0.84 & 0.81 & \underline{0.82} & 0.80 & 0.85 & 0.81 \\\hline
 & SS & 0.77 & 0.73 & 0.74 &  0.86 & 0.79 & \underline{0.81} \\
 & SI & 0.94 & 0.94 & \underline{0.94} &  0.82 & 0.82 & 0.82\\
 & Average & 0.86  & 0.85 & \underline{0.85} & 0.84 & 0.81 & 0.82  \\\hline
 
\end{tabular}
\end{table}

\subsection{System Performance}
Analysis was conducted using the gold-standard, manually annotated data. The macro-averaged precision, recall, and f-scores for classifying fine- and coarse-grained SS and SI categories at note level are provided in Table~\ref{main-result}. 

At MSHS, the RBS outperformed the LLM-based system for both fine- and coarse-grained classification. For the fine-grained categories, the RBS achieved macro-averaged f-score of 0.90 compared to 0.62 for the LLM. For coarse-grained classification, the RBS had macro-averaged f-score of 0.89 versus 0.65 for the LLM. 

At WCM, the RBS outperformed the LLM for fine-grained classification with macro-averaged f-scores of 0.82 versus 0.81, respectively. The coarse-grained categories were similar, with a macro-averaged f-score of 0.85 for the RBS compared to 0.82 for FLAN. The performance of the zero-shot FLAN-T5-XL model is provided in Supplementary Table~\ref{llm-result}.


\textbf{Comparison to ICD codes:} There were zero visits associated with the annotated clinical notes where SI was captured by the structured ICD codes (ICD-10: `Z60.2’, `V60.3’, `Z60.4’, `Z60.9’; ICD-9: `V60.3’, `V62.4’ [see Supplementary Table~\ref{tab:icd_definition}]). Without the NLP pipelines, the presence of SI would otherwise be missed both sites. 

\section{Discussion}

This study presents rule- and LLM-based NLP systems to identify fine-grained categories of SS and SI in clinical notes of psychiatric patients. A primary goal of the study was to develop and validate two portable and open-source NLP systems. Given that none of the selected clinical notes were associated with ICD codes indicating SI, the development of both NLP systems enabled the identification and subcategorization of this risk factor.

Comparative accuracy performance was initially unexpected given that LLMs typically outperform RBS on related tasks~\cite{lybarger2023advancements}. Upon manual review of the results, it became evident that the rule- and LLM-based approaches solved the task in different ways, both of which could be considered valid. However, these differences are not reflected in the performance metrics. The rule-based lexicon approach appears to have been significantly better than the LLM since it is most similar to the manual annotation rule book and thus, the gold-standard annotations that are assigned as ground-truth when evaluating system performance. The rule book and the lexicons were developed together, with the goal of the lexicon-based approach to approximate the Rules as closely as possible. 

Furthermore, the gold-standard annotations and the RBS assign a single label for each SI/SS occurrence, whereas the LLM system can assign multiple labels. This is a consequence of having separate fine-tuned LLMs for each of the SI and SS subcategories. Future work is warranted to improve model accuracy when adapting COT question-answering for multilabel classification tasks~\cite{guevara2023large}. Another difference is that the rule book and lexicons took a conservative approach, only assigning a label if the concept was explicit, whereas the LLM was more flexible. For  example, `\textit{She feels depressed and suicidal because she has no friends and no boyfriend}’ was labeled by the RBS as well as the gold-standard annotation as \textit{no social network} because having no friends is in the lexicon and rule book. In contrast, the LLM inferred both \textit{no social network} and \textit{loneliness}.

Comparing efficiency in the development of the RBS and LLMs, we assumed that the LLMs would require less manual input given that FLAN-T5 is ``few-shot,'' requiring no labeled training data. However, without fine-tuning, the model performed poorly and required synthetic examples. The iterative validation process revealed that strategic tuning examples were required to coerce the LLM to override the colloquial understanding of the categories for the task-specific definitions. Still, there were some concepts that the LLM would not unlearn during fine-tuning, even with a higher learning rate. For example, in the \textit{no instrumental support} model, the synthetic example `doesn’t have a lot of spending money on hand to engage in the activities she would like’ continued to be labeled `yes’ rather than the correct `not relevant’ label. There were many cases where the LLM identified the presence of SI/SS correctly that were neither identified by the RBS nor present in the gold-standard annotations as dictated by the rule book. For example, `\textit{Lived in an Assisted Living facility for a year},' and `\textit{Pt hasn't been in touch with her family}’ (see Supplemental Table~\ref{tab:examples_both} for more examples).


The performances of both the RBS and LLMs were relatively poor in identifying \textit{instrumental support}. This is likely because the keywords often contain site-specific names and entities such as `\textit{HASA}' and `\textit{Lenox Hill Neighborhood House},' to name a couple of the many examples. Another area where both approaches fell short was in the \textit{plan} section of the psychiatric clinical note. Part of the clinical plan might be to increase social connectedness. The manual annotators were easily able to understand that context whereas both systems generated false positive labels. Further erroneous examples from the rule- and LLM-based systems are provided in Supplementary Table~\ref{tab:examples_both}.

The RBS performed comparably at MSHS and WCM; however, the LLM performed better at WCM compared to MSHS. This is likely related to two key differences. The first is the higher frequency of SS/SI mentions at MSHS (e.g., 75.3\% of notes at MSHS had a manually annotated mention of SS vs. 52.2\% at WCM). The second, related difference is that, in addition to SS/SI mentions that fit the inclusion criteria for manual annotations, MSHS also has more mentions of SS/SI concepts that did not fit within the strict rule book of the manual annotations nor the lexicons of the RBS, but were identified by the LLM. This is due to the MSHS underlying corpus being from clinical care sites (such as inpatient psychiatry) with comprehensive psychiatric evaluations that systematically include SDOH information. 

This work expands on the body of literature by specifically focusing on the fine-grained classification of SS and SI, a novel approach not undertaken by earlier studies. More broadly, Guevara et al.~\cite{guevara2023large} utilized LLM-based classification for SS and adverse SS (SI). Their study reported the best f-scores of 0.60 (FLAN-T5-XXL) and 0.56 (FLAN-T5-XL) for SS and adverse SS, respectively, across 154 test documents. In contrast, Zhu et al.~\cite{zhu2019automatically} deployed 24 lexicons and the Linguamatics I2E NLP tool for identifying SI and achieving an f-score of 0.93 from 194 clinical notes. Our study presents superior outcomes across two sites and with more specific (fine-grained) categories.

\subsection{Limitations}

Several limitations should be noted. There were insufficient instances in the notes of the emotional support subcategories to evaluate the NLP systems. Emotional support (and lack thereof) is an important and distinct fine-grained category that would ideally be identified in the notes. Second, the RBS was designed with specific lexicons from manual review at MSHS and WCM, may have experienced overfitting and led to an inflated f-score. It would be beneficial to validate these NLP systems on clinical notes from different EHR systems. Other healthcare systems that implement a lexicon-based rules approach will need to perform site-specific template removal to avoid the problem of false-positives. With fine-tuning, the LLM approach may have been able to correctly interpret the templates; however, because the templates were removed from the notes before the annotation process, this was not assessed.

\section{Conclusion}
We offer two open-source NLP systems with different approaches, as well as a manual annotation guideline for identifying SS and SI. The rule-based approach and the LLM approach each have strengths and limitations in performing this challenging task of creating a portable system to identify fine and coarse-grained categories of SS and SI in psychiatric patients. 

\section*{Reproducibility}
All code for rule- and LLM-based systems are available on GitHub.  \\\textbf{Link:} \href{https://github.com/CornellMHILab/Social\_Support\_Social\_Isolation\_Extraction}{https://github.com/CornellMHILab/Social\_Support\_Social\_Isolation\_Extraction}.\\
The annotation guidelines and lexicons are available in the supplementary files.

\section*{Funding}
This study was funded in part by grants from the National Institutes of Health (R01 MH119177, R01 MH121907, and R01 MH121922).

\section*{Acknowledgements}
This work was supported in part through the computational and data resources and staff expertise provided by Scientific Computing and Data at the Icahn School of Medicine at Mount Sinai Health System and supported by the Clinical and Translational Science Awards (CTSA) grant UL1TR004419 from the National Center for Advancing Translational Sciences. Research reported in this publication was also supported by the Office of Research Infrastructure of the National Institutes of Health under award number S10OD026880 and S10OD030463. The content is solely the responsibility of the authors and does not necessarily represent the official views of the National Institutes of Health.

\section*{Author Contributions}
BGP and LAL worked on planning, data collection, data annotation, data extraction, software development, writing, and editing. PR worked on LLM implementation and editing. VV helped in data annotation, writing, and editing. MMS, BF, and GH helped in data annotation. JP and PA helped in planning, advising, and editing. MW, JJM, and MO helped in initial category identification, lexicon development, advising, and editing. JASR, ER, JMB, IL, GNN, AT, and AWC helped in advising and editing. JP and AWC provided resources and funding.

\newpage
\section*{Supplementary}
\beginsupplement

\section*{Guidelines for annotating social support \\and social isolation in clinical notes}
\label{sup_ann_guide}

\noindent Social isolation (SI) is a state of little or no contact or support between an individual and their community. Social support (SS) describes the physical and emotional comfort derived from one's family, friends, colleagues, and community. Although electronic health records (EHRs) capture SI and SS within the narrative text, manual abstraction of such data is time-consuming and labor-intensive.

\subsection*{Research objective}
To develop a hybrid approach using rule-based and machine learning (ML)-based natural language processing (NLP) for detecting patient-reported SS and SI in EHRs. 

\subsection*{Annotation task overview}
The purpose of annotating clinical notes is to highlight key terms and phrases that characterize one's social standing and, based on their context, make a clinical determination of an individual's present or past status with respect to SI and SS.

\subsection*{Annotation process}

\textbf{Guideline development phase:} Initially, two annotators will annotate fifty (50) notes. Then, inter-annotator agreement (IAA) will be calculated, and the annotation guidelines will be revised. This process will continue until a sufficiently high agreement (IAA $\geq$ 0.6) is achieved. The mismatched cases will be adjudicated by iterative discussion until a consensus is reached. The first 50 clinical notes will be added to the gold standard but excluded from the final IAA computation. 

\textbf{Annotation Sites:} Weill Cornell Medicine (WCM) and Mount Sinai Health System (MSHS).


\textbf{Annotation Phase:} Each document will be independently annotated by two annotators. After the annotations are completed, the conflicts will be resolved by a third annotator. The annotation guideline will be updated if needed. Finally, the IAA will be calculated on all documents excluding the first 50 documents that are used to improve the annotation and guideline. 

\textbf{Notes:} MSHS will annotate psychiatry notes only as they don't have access to primary care notes right now. In the annotation cohort, a single patient will have only one note. Entity attribute ``negation'' will be used only for ``social\_isolation\_loneliness'' category. 

\subsection*{General guidelines for annotation}

Annotators will annotate all mentions in a clinical note that indicate the presence (or absence) of present/past SI and SS. There may be multiple such mentions contained within a single document. The goal is to make an aggregate determination based on the various annotations within the document. The overall annotation schema is given in Table~\ref{schema_annotation}. The annotators will annotate all terms and phrases mentioned in Table~\ref{lexicons} that indicate an individual's social status based on the annotation schema. 
A single document may contain multiple entities from both SI and SS. Further, the annotators will annotate at the document/note level based on entities and their context available in that specific note. There are two tags for annotating at the document level and these are mentioned in Table~\ref{schema_annotation}.



\begin{table}[]
\caption{Annotation schema for social isolation and support}
\label{schema_annotation}
\begin{tabular}{llll}
\hline
\multicolumn{4}{l}{\textbf{Entity Level}} \\\hline

\textbf{Social Isolation} & \textbf{Social Support} & \textbf{Temporality} & \textbf{Negation} \\ \hline

loneliness & -- & Present  & Yes \\

no social network & social network & Past &  \\

no emotional support & emotional support & &  \\

no instrumental support & instrumental support & &  \\ 
general & general & & \\
probable & probable & & \\ \hline
\multicolumn{4}{l}{\textbf{Document/Note Level}} \\\hline
\multicolumn{4}{l}{social isolation (yes/no)}\\
\multicolumn{4}{l}{social support (yes/no)} \\ \hline
\end{tabular}

\end{table}

\subsection*{Instructions for annotation}
Annotations will be performed using a BRAT annotation tool made available at individual sites. This is a browser-based tool that users may access online in a secure environment. Each user will be provided a username and password to access the tool. Before starting the actual annotation process, each user will participate in an orientation session to familiarize themselves with the BRAT tool. Within the tool, annotators will be granted access to 300 [MS] and 225 [WCM] clinical notes provided as .txt files. After opening a .txt file, annotators will highlight spans of pertinent text and assign tags based on the entities in Table~\ref{schema}. In some cases, the highlighted terms may be used in an unrelated context (e.g., templated information: \textit{social isolation []}) and can be tagged with `no\_context.'

We plan to annotate \textit{social isolation} and \textit{social support} at a fine-grained entity level. There are five (5) categories for social isolation: ``\textit{loneliness},'' ``\textit{no social network},'' ``\textit{no emotional support},'' ``\textit{no instrumental support},'' and ``\textit{social isolation general}." Similarly, social support has four (4) categories: ``\textit{social network},'' ``\textit{emotional support},'' ``\textit{instrumental support},'' and ``\textit{general}." Each term can be annotated with temporal information (present or past) and with negation (if a negation term precedes or follows the highlighted key terms [only for loneliness]). There are two (2) categories for document/note level annotation and these are ``\textit{social isolation}'' and ``\textit{social support}'' with ``\textit{yes}" and ``\textit{no}". The document/note level annotations are provided based on the entity level annotations and the rules are provided in Table~\ref{tab:rules_annotation}. Table~\ref{lexicons} shows the lists of key terms and phrases that we have used for the development of NLP algorithm. We also identified some key terms and phrases that were excluded while identifying the social support or isolation from clinical encounter notes (See Table~\ref{exclusion_terms}).   

\begin{table}[!htbp]
    \centering
    \caption{Rules for identifying coarse-grained categories based on the fine-grained categories.}
    \begin{tabular}{l|l}
        \hline
         \textbf{Categories} & \textbf{Rules}  \\ \hline
         \textbf{SS} & if there are one or more mentions of any SS subcategories\\ \hline
        \textbf{SI} & if there are one or more mentions of any SI subcategories\\ \hline
        \textbf{None} & if there is no mention of either SI/SS subcategory\\ \hline
    \end{tabular}
    
    \label{tab:rules_annotation}
\end{table}

\newpage

\begin{longtable}{| p{.98\textwidth} | p{.98\textwidth} |} 
\hline
\textbf{Social isolation: loneliness}  \\ \hline
alone in the world \\ \hline
aloneness \\ \hline
experiencing feelings of isolation \\ \hline
feel isolate \\ \hline
feel isolated \\ \hline
feeling alone \\ \hline
feeling isolated \\ \hline
feeling of not having anyone \\ \hline
feelings of isolation \\ \hline
feels all alone \\ \hline
feels alone \\ \hline
feels isolate \\ \hline
feels isolated \\ \hline
loneliness \\ \hline
lonely \\ \hline
sense of isolation \\ \hline
\textbf{Social isolation: no social network}  \\ \hline
denies having any friends \\ \hline
denies having close friends \\ \hline
denies having friends \\ \hline
does not have a support network \\ \hline
does not have any close family or friends \\ \hline
does not have any close friends \\ \hline
does not have any family or close friends \\ \hline
does not have any family or friends \\ \hline
does not have any friends \\ \hline
does not have any friends or family \\ \hline
does not have close family or any friends \\ \hline
does not have close friends \\ \hline
does not have family \\ \hline
does not have family or friends \\ \hline
does not have friends \\ \hline
does not have friends or family \\ \hline
does not have many close friends \\ \hline
does not have many friends \\ \hline
does not socialize \\ \hline
doesn't have family or friends \\ \hline
few friends \\ \hline
few social contacts \\ \hline
had no family \\ \hline
had no friend \\ \hline
had no friends \\ \hline
has 0 friends \\ \hline
has no family \\ \hline
has no friend \\ \hline
has no friends \\ \hline
has no social activities \\ \hline
having no friend \\ \hline
increase social activity \\ \hline
lack of a support network \\ \hline
lack of support network \\ \hline
limited social connection \\ \hline
limited social network \\ \hline
little social contact \\ \hline
loss of social network \\ \hline
no close family or friends \\ \hline
no close friend \\ \hline
no close friends \\ \hline
no family \\ \hline
no family and friends \\ \hline
no family or friends \\ \hline
no friend \\ \hline
no friends \\ \hline
no friends and family \\ \hline
no one to socialize \\ \hline
no relatives \\ \hline
no social contact \\ \hline
no social network \\ \hline
no socialization \\ \hline
no support network \\ \hline
not yet have friends \\ \hline
poor social network \\ \hline
reports no friend \\ \hline
reports no friends \\ \hline
socially disconnected \\ \hline
wants to have friends \\ \hline
wants to make friends \\ \hline
wants to make new friends \\ \hline
\textbf{Social isolation: no emotional support}  \\ \hline
absence of emotional support \\ \hline
can talk to no one \\ \hline
can talk to nobody \\ \hline
can't confide in \\ \hline
can't get emotional support \\ \hline
can't get the emotional support \\ \hline
can't speak openly \\ \hline
can't talk to anybody \\ \hline
can't talk to anyone \\ \hline
cannot confide in \\ \hline
cannot speak openly \\ \hline
cannot talk to anybody \\ \hline
cannot talk to anyone \\ \hline
deficit in emotional support \\ \hline
does not confide in \\ \hline
does not feel comfortable sharing \\ \hline
does not feel comfortable sharing personal \\ \hline
does not have emotional support \\ \hline
doesn't confide in \\ \hline
doesn't feel comfortable sharing \\ \hline
doesn't have emotional support \\ \hline
in need of emotional support \\ \hline
isn't comfortable sharing \\ \hline
lacking emotional support \\ \hline
lacks emotional support \\ \hline
limited emotional support \\ \hline
needs emotional support \\ \hline
no emotional support \\ \hline
no one for emotional support \\ \hline
no one to confide in \\ \hline
no one to share personal \\ \hline
no one to speak openly \\ \hline
no one to talk to \\ \hline
no one understands \\ \hline
no one who truly understands \\ \hline
nobody cares \\ \hline
nobody to share personal \\ \hline
not comfortable sharing \\ \hline
not emotional support \\ \hline
prevents her from confiding \\ \hline
prevents him from confiding \\ \hline
reluctance to confide in others \\ \hline
talk to no one \\ \hline
to increase emotional support \\ \hline
unable to confide in \\ \hline
will not confide in \\ \hline
won't confide in \\ \hline
would not feel comfortable sharing personal \\ \hline
\textbf{Social isolation: no instrumental support}  \\ \hline
did not have medicaid \\ \hline
does not have hha \\ \hline
does not have medicaid \\ \hline
homeless \\ \hline
homelessness \\ \hline
lives on the street \\ \hline
living on the street \\ \hline
losing her housing \\ \hline
losing his housing \\ \hline
lost her housing \\ \hline
lost his housing \\ \hline
needs assistance with adls \\ \hline
needs assistance with iadls \\ \hline
needs help with adls \\ \hline
needs hha \\ \hline
no hha \\ \hline
no home care services \\ \hline
no homecare services \\ \hline
no one to help with chores \\ \hline
no one to help with daily chores \\ \hline
no one to help with housework \\ \hline
no one to help with iadls \\ \hline
requires assistance with adls \\ \hline
sleeps on the streets \\ \hline
street homeless \\ \hline
undomicilied \\ \hline
\textbf{Social isolation: general}  \\ \hline
address isolation \\ \hline
all alone \\ \hline
appears to have limited social support \\ \hline
chronic isolation \\ \hline
combating her isolation \\ \hline
combating his isolation \\ \hline
dealing with isolation \\ \hline
denies having any other family/support \\ \hline
denies having any other support \\ \hline
denies having any support \\ \hline
denies having family or friend support \\ \hline
denies having family support \\ \hline
denies having friend or family support \\ \hline
describes limited social support \\ \hline
does not identify any social support \\ \hline
does not participate in social activities \\ \hline
has limited social support \\ \hline
has little to no social support \\ \hline
has no family support \\ \hline
has no social support \\ \hline
increasingly isolated \\ \hline
is alone \\ \hline
isolate herself \\ \hline
isolate himself \\ \hline
isolated from everyone \\ \hline
isolated from peers \\ \hline
isolated with limited social support \\ \hline
isolates herself \\ \hline
isolates himself \\ \hline
isolating at home \\ \hline
isolating herself \\ \hline
isolating himself \\ \hline
lack in social support \\ \hline
lack in social supports \\ \hline
lack of psychosocial support \\ \hline
lack of social and family support \\ \hline
lack of social support \\ \hline
lack of social supports \\ \hline
life of isolation \\ \hline
limited family support \\ \hline
limited social connection \\ \hline
limited social support \\ \hline
limited social supports \\ \hline
limited social/family support \\ \hline
limited support system \\ \hline
limited support systems \\ \hline
limited supports system \\ \hline
little family or social support \\ \hline
little family support \\ \hline
little social support \\ \hline
little to no family support \\ \hline
no family support \\ \hline
no family supports \\ \hline
no family/friend supports \\ \hline
no on in her family supports her \\ \hline
no one in his family supports him \\ \hline
no social interaction \\ \hline
no social support \\ \hline
no social supports \\ \hline
poor social support \\ \hline
poor to no social support \\ \hline
reduce her isolation \\ \hline
reduce his isolation \\ \hline
reported having no social support \\ \hline
reports having no social support \\ \hline
reports limited social support \\ \hline
reports no social support \\ \hline
social contact limited \\ \hline
social isolation \\ \hline
social withdraw \\ \hline
social withdrawal \\ \hline
social withdrawal behavior \\ \hline
social withdrawal behaviors \\ \hline
socially disconnected \\ \hline
socially isolated \\ \hline
socially isolating \\ \hline
socially withdrawn \\ \hline
tendency to isolate \\ \hline
very limited social support \\ \hline
with isolation \\ \hline
\textbf{Social support: social network}  \\ \hline
active in her church \\ \hline
active in her mosque \\ \hline
active in her religious \\ \hline
active in her synagogue \\ \hline
active in her temple \\ \hline
active in his church \\ \hline
active in his mosque \\ \hline
active in his religious \\ \hline
active in his synagogue \\ \hline
active in his temple \\ \hline
attending church \\ \hline
attends church \\ \hline
close friend \\ \hline
close friends \\ \hline
close relationship with \\ \hline
club house \\ \hline
commune \\ \hline
community activities \\ \hline
community activity \\ \hline
domiciled with friend \\ \hline
domiciled with friends \\ \hline
family connection \\ \hline
family meeting \\ \hline
family visit \\ \hline
friend connection \\ \hline
friend of a friend \\ \hline
goes to church \\ \hline
going to church \\ \hline
has a good friend \\ \hline
has brother \\ \hline
has close friend \\ \hline
has close friends \\ \hline
has family \\ \hline
has friend \\ \hline
has friends \\ \hline
has good friends \\ \hline
has lots of friends \\ \hline
has many friends \\ \hline
has sister \\ \hline
has special person \\ \hline
have a good friend \\ \hline
have bother \\ \hline
have close friend \\ \hline
have close friends \\ \hline
have family \\ \hline
have friend \\ \hline
have friends \\ \hline
have good friends \\ \hline
have many friends \\ \hline
have sister \\ \hline
have special person \\ \hline
increased social activity \\ \hline
increased socialization \\ \hline
is popular \\ \hline
lives with friend \\ \hline
lives with friends \\ \hline
moved in with friends \\ \hline
mutual friends \\ \hline
reported having friends \\ \hline
reports having friends \\ \hline
see friends \\ \hline
sees friend \\ \hline
sees friends \\ \hline
social activity \\ \hline
social network \\ \hline
social support network \\ \hline
staying at a friend's \\ \hline
staying at friend's \\ \hline
support network \\ \hline
visit church \\ \hline
visit family \\ \hline
visits church \\ \hline
visits family \\ \hline
was popular \\ \hline
with friend \\ \hline
with friends \\ \hline
\textbf{Social support: emotional support}  \\ \hline
brother shares \\ \hline
brother understands \\ \hline
brothers share \\ \hline
brothers understand \\ \hline
can talk about her problem \\ \hline
can talk about his problem \\ \hline
can talk about my problem \\ \hline
can talk to \\ \hline
comfortable sharing \\ \hline
confide in \\ \hline
emotional support \\ \hline
father shares \\ \hline
father understands \\ \hline
feeling understood \\ \hline
feelings are openly discussed \\ \hline
feels comforted \\ \hline
feels heard \\ \hline
feels understood \\ \hline
friend shares \\ \hline
friend understands \\ \hline
friends share \\ \hline
friends understand \\ \hline
grand father shares \\ \hline
grand father understands \\ \hline
grand mother shares \\ \hline
grand mother understands \\ \hline
grand parent shares \\ \hline
grand parent understands \\ \hline
grand parents share \\ \hline
grand parents understand \\ \hline
mother shares \\ \hline
mother understands \\ \hline
openly discuss her feelings \\ \hline
openly discuss his feelings \\ \hline
openly discusses her feelings \\ \hline
openly discusses his feelings \\ \hline
parent shares \\ \hline
parent understands \\ \hline
parents share \\ \hline
parents understand \\ \hline
share experiences \\ \hline
share personal \\ \hline
shares experiences \\ \hline
shares personal \\ \hline
sister shares \\ \hline
sister understands \\ \hline
sisters share \\ \hline
sisters understand \\ \hline
speak openly \\ \hline
trust my brother \\ \hline
trust my father \\ \hline
trust my friend \\ \hline
trust my grand father \\ \hline
trust my grand mother \\ \hline
trust my grand parents \\ \hline
trust my mother \\ \hline
trust my parent \\ \hline
trust my sister \\ \hline
trusts her brother \\ \hline
trusts her father \\ \hline
trusts her friend \\ \hline
trusts her grand father \\ \hline
trusts her grand mother \\ \hline
trusts her grand parents \\ \hline
trusts her mother \\ \hline
trusts her parent \\ \hline
trusts her sister \\ \hline
trusts his brother \\ \hline
trusts his father \\ \hline
trusts his friend \\ \hline
trusts his grand father \\ \hline
trusts his grand mother \\ \hline
trusts his grand parents \\ \hline
trusts his mother \\ \hline
trusts his parent \\ \hline
trusts his sister \\ \hline
understands what he is going through \\ \hline
understands what she is going through \\ \hline
\textbf{Social support: instrumental support}  \\ \hline
assistant through hasa \\ \hline
assists with adls \\ \hline
at shelter \\ \hline
brother drives \\ \hline
brothers drive \\ \hline
domiciled in a nursing home \\ \hline
domiciled in a shelter \\ \hline
domiciled in shelter \\ \hline
father drives \\ \hline
friend drives \\ \hline
friends drive \\ \hline
has a cleaning lady \\ \hline
has hha \\ \hline
has housekeeper \\ \hline
has housekeeping help \\ \hline
has social service \\ \hline
have a cleaning lady \\ \hline
have hha \\ \hline
have housekeeper \\ \hline
have medicaid \\ \hline
have social service \\ \hline
having hha \\ \hline
having medicaid \\ \hline
having social service \\ \hline
help with cleaning \\ \hline
help with cooking \\ \hline
help with shopping \\ \hline
helps with adls \\ \hline
helps with cleaning \\ \hline
helps with cooking \\ \hline
helps with shopping \\ \hline
home attendant \\ \hline
home care \\ \hline
home health aide \\ \hline
homecare services \\ \hline
homeless shelter \\ \hline
lives in a nursing \\ \hline
lives in a shelter \\ \hline
lives in nh \\ \hline
lives in shelter \\ \hline
living at halfway house \\ \hline
living at shelter \\ \hline
living in a nursing \\ \hline
living in shelter \\ \hline
mother drives \\ \hline
on disability \\ \hline
parent drives \\ \hline
parents drive \\ \hline
part-time hha \\ \hline
sister drives \\ \hline
sisters drive \\ \hline
ssd \\ \hline
supported financially \\ \hline
supportive housing \\ \hline
transitional housing \\ \hline
with hha \\ \hline
\textbf{Social support: general}  \\ \hline
brother helps \\ \hline
brother supports \\ \hline
brothers help \\ \hline
brothers support \\ \hline
close family ties \\ \hline
close relationships \\ \hline
close with her \\ \hline
close with his \\ \hline
commune \\ \hline
family helping \\ \hline
family helps \\ \hline
family member helps \\ \hline
family members help \\ \hline
family members support \\ \hline
family support \\ \hline
family supportive \\ \hline
family supports \\ \hline
father helps \\ \hline
father supports \\ \hline
friend helps \\ \hline
friend supports \\ \hline
friends help \\ \hline
friends support \\ \hline
good family \\ \hline
grand father helps \\ \hline
grand father supports \\ \hline
grand mother helps \\ \hline
grand mother supports \\ \hline
grand parent helps \\ \hline
grand parent supports \\ \hline
grand parents help \\ \hline
grand parents support \\ \hline
has family for support \\ \hline
has family support \\ \hline
has help from family \\ \hline
less socially isolated \\ \hline
mother helps \\ \hline
mother supports \\ \hline
parent helps \\ \hline
parents help \\ \hline
she helps \\ \hline
sister helps \\ \hline
sister supports \\ \hline
sisters help \\ \hline
sisters support \\ \hline
social support \\ \hline
source of support \\ \hline
support from family \\ \hline
supportive family \\ \hline
supportive friends \\ \hline
supportive throughout \\ \hline
\caption{List of typical examples of terms and phrases in the clinical encounter notes that would indicate a patient's social support and isolation status.\label{lexicons}}

\end{longtable}

\begin{longtable}{| p{.98\textwidth} | p{.98\textwidth} |} 
\hline
\textbf{Exclusion Terms}  \\ \hline
$<$social\_isolation$<$social\_support \\ \hline
[-] social support \\ \hline
[-]social support \\ \hline
[] close relationships \\ \hline
[] social isolation \\ \hline
[] social support \\ \hline
[]close relationships \\ \hline
[]social support \\ \hline
any social support \\ \hline
applied for supportive housing \\ \hline
applying for supportive housing \\ \hline
asking for home care \\ \hline
asking for supportive housing \\ \hline
assessed for home care \\ \hline
assessed for supportive housing \\ \hline
benefit from home care \\ \hline
can talk to attending \\ \hline
can talk to billing \\ \hline
can talk to boss \\ \hline
can talk to case worker \\ \hline
can talk to doctor \\ \hline
can talk to front desk \\ \hline
can talk to hospital \\ \hline
can talk to nurse \\ \hline
can talk to provider \\ \hline
can talk to reap \\ \hline
can talk to social work \\ \hline
can talk to staff \\ \hline
can talk to sw \\ \hline
can talk to team \\ \hline
can talk to writer \\ \hline
cannot trust \\ \hline
can’t trust \\ \hline
current living conditions/support network \\ \hline
desire to have friends \\ \hline
desires hha \\ \hline
desires to have friends \\ \hline
does not trust \\ \hline
doesn’t trust \\ \hline
evaluated for supportive housing \\ \hline
has been hha \\ \hline
has been home health aid \\ \hline
has been home health aide \\ \hline
has worked as home health \\ \hline
in need of hha \\ \hline
in need of home health \\ \hline
inability to go to shelter \\ \hline
inability to make friends \\ \hline
increasing her social support \\ \hline
increasing his social support \\ \hline
lost supportive housing \\ \hline
needing home care \\ \hline
needs hha \\ \hline
needs home care \\ \hline
needs home health aid \\ \hline
needs home health aide \\ \hline
needs supportive housing \\ \hline
no family available \\ \hline
no family history \\ \hline
no family hx \\ \hline
no trust \\ \hline
not married \\ \hline
paperwork for supportive housing \\ \hline
parents are divorced \\ \hline
parents are getting divorced \\ \hline
parents divorced \\ \hline
parents getting divorced \\ \hline
parents got divorced \\ \hline
partner status \\ \hline
qualify for home care \\ \hline
qualify for supportive housing \\ \hline
social isolation? \\ \hline
they are getting divorced \\ \hline
trust her body \\ \hline
trust her doctor \\ \hline
trust her intuition \\ \hline
trust her nurse \\ \hline
trust her provider \\ \hline
trust her self \\ \hline
trust her team \\ \hline
trust her treatment \\ \hline
trust herself \\ \hline
trust him self \\ \hline
trust himself \\ \hline
trust his body \\ \hline
trust his doctor \\ \hline
trust his intuition \\ \hline
trust his nurse \\ \hline
trust his provider \\ \hline
trust his team \\ \hline
trust his treatment \\ \hline
trust my body \\ \hline
trust my education \\ \hline
trust my intuition \\ \hline
trust my self \\ \hline
trust myself \\ \hline
wants home care \\ \hline
wants supportive housing \\ \hline
works as hha \\ \hline
works as home health \\ \hline
would benefit from supportive housing \\ \hline
\caption{List of typical examples of exclusion terms and phrases in the clinical notes.\label{exclusion_terms}}
\end{longtable}

\subsection*{File naming convention}

Documents are named in the following naming convention: \\
$<$Index number$>$\_$<$person\_id$>$\_$<$note\_id$>$\_$<$note\_date$>$.txt\\
E.g.: 1\_101\_1001\_10001\_100001\_2000-01-01.txt

\noindent\textbf{References}

\begin{enumerate}
    \item Lauren A. Lepow, Braja Gopal Patra, et al. 2021. Extracting Social Isolation Information From Psychiatric Notes in the Electronic Health Records. In \textit{AMIA 2021 Annual Symposium}, San Diego, CA, USA.
\end{enumerate}

\newpage

\begin{table}[!htbp]
    \centering
    \caption{Document-level counts of different fine- and coarse-grained categories at MSHS and WCM.}
    \label{tab:iaa:dataset}
    \begin{tabular}{cl|cc}\hline
    \multicolumn{2}{c|}{\textbf{Categories}} & \multicolumn{2}{c}{\textbf{Counts (\%)}}\\ \hline
    & & \textbf{WCM} & \textbf{MSHS} \\ \hline
    \multicolumn{4}{c}{\textbf{Fine Grained Annotation}} \\ \hline
    \multirow{5}{*}{\textbf{SS}} & Soc. Network & ~81 (35.2\%) & ~76 (25.3\%) \\
    & Emotional & ~23 (10.0\%) & ~12 ~(4.0\%)\\
    & Instrumental & 102 (44.4\%) & ~61 (20.3\%) \\
    & general & ~58 (25.2\%) & 114 (38.0\%) \\
\hdashline
    \multirow{6}{*}{\textbf{SI}}& loneliness & ~33 (14.4\%) & ~21 ~(7.0\%)\\
    & no Soc. Network & ~50 (21.7\%) & ~25 ~(8.3\%)\\
    & no Emotional & ~12 ~(5.2\%) & ~~4 ~(1.3\%) \\
    & no Instrumental & ~56 (24.4\%)& ~56 (18.7\%)\\
    & general & ~76 (33.0\%)  & ~70 (23.3\%)\\\hline
    \multicolumn{4}{c}{\textbf{Coarse Grained Annotation}} \\ \hline
    & SS & 143 (52.2\%) & 226 (75.3\%) \\ 
    & SI & 122 (53.0\%) & 127 (42.3\%)\\ \hline
    \end{tabular}
\end{table}

\newpage

\section*{Other Supervised Models}

We performed a sentence-level classification using SVM and BERT. The entity-level annotation were converted to sentence-level. 

\textbf{SVM:} Initially, we used two VSM methods namely term frequency-inverse document frequency (TF-IDF) and word2vec [1] to convert the clinical notes into vectors. However, the word2vec-based system performed better than the TF-IDF-based system and we reported the performance of word2vec-based system next. We used the same embeddings that we used to identify similar words. We used Eq~\ref{eq1} to convert the word vectors to a single vector. The embeddings settings are as follows: vector size: 300, minimum word count: 3, and window size: 10. We used the linear kernel in the SVM classifier. 

\textbf{BERT:} We leveraged the state-of-the-art BERT [2] and further fine-tuned the pre-trained model in downstream tasks. The model took a sequence of tokens with a maximum length of 512 and produces a 768-dimensional sequence representation vector. For text that is shorter than 512 tokens, we added paddings (empty tokens) to the end of the text to make up the length. For text that is longer than 512 tokens, we used the first 512 tokens as the input. Then, two fully connected layers are appended on top of the pooler output layer of the BERT model. Finally, a SoftMax layer is used to map the representation vector to the target label space. For the BERT model, we fine-tuned the `BioBERT' [3] model with the training data for 20 epochs with a learning rate of $2\times10^{-5}$ and batch size of 16. We adopted AdamW [4] as the optimizer and cross-entropy as the loss function. 

\begin{table}[!htbp]
    \centering
    \caption{Macro-averaged precision (P), recall (R) and F-scores (F) of SVM- and BERT-based NLP systems for coarse-categories on data from WCM only. Here, we split the data in 80:20 ratio for training and testing.}
    \label{tab:mldl_results}
    \begin{tabular}{c|ccc|ccc}
    \hline
    \textbf{Categories} & \multicolumn{3}{|c|}{\textbf{SVM}} & \multicolumn{3}{c}{\textbf{BERT}} \\ \cline{2-7}
    & \textbf{P} & \textbf{R} & \textbf{F} & \textbf{P} & \textbf{R} & \textbf{F} \\\hline
    SS & 0.55 & 0.71 & 0.62 & 0.59 & 0.59 & 0.59  \\
    SI & 0.68 & 0.58 & 0.62 & 0.62 & 0.62 & 0.62 \\
    Average & 0.61 & 0.58 & 0.60 & 0.61 & 0.60 & 0.61\\\hline
    \end{tabular}
    
\end{table}

\subsection*{References}

\begin{enumerate}
    \item Tomas Mikolov, Ilya Sutskever, Kai Chen, Greg S Corrado, and Jeff Dean. Distributed representations of words and phrases and their compositionality. Advances in neural information processing systems, 26, 2013.
    \item Jacob Devlin, Ming-Wei Chang, Kenton Lee, and Kristina Toutanova. Bert: Pre-training of deep bidirectional transformers for language understanding. arXiv preprint arXiv:1810.04805, 2018.
    \item Jinhyuk Lee, Wonjin Yoon, Sungdong Kim, Donghyeon Kim, Sunkyu Kim, Chan Ho So, and Jaewoo Kang. Biobert: a pre-trained biomedical language representation model for biomedical text mining. Bioinformatics, 36(4):1234–1240, 2020.
    \item Diederik P Kingma and Jimmy Ba. Adam: A method for stochastic optimization. arXiv preprint arXiv:1412.6980, 2014.

\end{enumerate}

\newpage

\begin{table}[!htbp]
    \centering
    \caption{Questions for fine-tuning LLM model for different categories.}
    \label{questions}
    \begin{tabular}{p{3.5cm}|p{9cm}}
    \hline
    \textbf{Category} & \textbf{Question} \\ \hline
    \multicolumn{2}{l}{\textbf{Social Isolation}} \\ \hline
    loneliness & In the clinician's opinion, does or did the patient experience feelings of loneliness? \\\hline
    no social network & In the clinician's opinion, does or did the patient lack a social network? \\\hline
    no emotional support & In the clinician's opinion, does or did the patient lack emotional support? \\\hline
    no instrumental support & In the clinician's opinion, does or did the patient lack access to instrumental support? \\ \hline
    general & In the clinician's opinion, does or did the patient have social isolation; however, there is not enough information to classify the type of social isolation as specifically lack of instrumental support, lack of emotional support, lack of social network, or loneliness?  \\ \hline
    \multicolumn{2}{l}{\textbf{Social Support}}\\ \hline
    social network & In the clinician's opinion, does or did the patient have a social network? \\\hline
    emotional support & In the clinician's opinion, does or did the patient have adequate emotional support?\\\hline
    instrumental support & In the clinician's opinion, does or did the patient have access to instrumental support? \\\hline
    general & In the clinician's opinion, does or did the patient have social support; however, there is not enough information to classify the type of social support as specifically instrumental support, emotional support, or social network?\\
    \hline
    \end{tabular}
\end{table}

\begin{table}[!htbp]
\centering
\caption{Macro-averaged Precision (P), Recall (R), and F-scores (F) for fine and coarse-grained category classification using FLAN-T5-XL. Here, we used the instructions and did not fine-tuning the model using the examples.}
\label{llm-result}
\begin{tabular}{cl|ccc}
\hline
\multicolumn{2}{c}{\multirow{2}{*}{\textbf{Categories}}} & \multicolumn{3}{|c}{\textbf{FLAN-T5-XL}} \\ \cline{3-5}
& & \textbf{P} & \textbf{R} & \textbf{F} \\\hline
\multicolumn{5}{c}{\textbf{WCM}} \\\hline
\multirow{4}{*}{\textbf{SS}} & Soc. Network & 0.67 & 0.61 & 0.47 \\
 & Emotional & 0.55 & 0.68 & 0.48 \\
 & Instrumental & 0.62 & 0.54 & 0.43 \\
 & general & 0.62 & 0.51 & 0.21 \\ \hdashline
\multirow{5}{*}{\textbf{SI}} & loneliness & 0.60 & 0.66 & 0.41 \\
 & no Soc. Network & 0.67 & 0.70 & 0.54 \\
 & no Emotional & 0.53 & 0.63 & 0.27 \\
 & no Instrumental & 0.64 & 0.61 & 0.41 \\
 & general & 0.66 & 0.51 &  0.27\\
 & average & 0.63 & 0.62 & 0.43 \\\hline
 & SS & 0.30 & 0.50 & 0.37 \\
 & SI & 0.79 & 0.52 & 0.39 \\
 & average & 0.67 & 0.51 &  0.38  \\\hline
\multicolumn{5}{c}{\textbf{MSHS}} \\\hline 
\multirow{4}{*}{\textbf{SS}} & Soc. Network & 0.68 & 0.58 & 0.42 \\
 & Emotional  & 0.57 & 0.56 & 0.57  \\
 & Instrumental &  0.56 & 0.51 & 0.21 \\
 & general & 0.64 & 0.56 & 0.50 \\\hdashline
\multicolumn{1}{c}{\multirow{5}{*}{\textbf{SI}}} & loneliness & 0.55 & 0.53 & 0.15 \\
 & no Soc. Network & 0.55 & 0.54 & 0.17\\
 & no Emotional & 0.51 & 0.53 & 0.07   \\
& no Instrumental & 0.74 & 0.57 & 0.45   \\
& general & 0.79 & 0.72 & 0.68 \\
 & average & 0.56 & 0.53 & 0.36    \\\hline

 & SS & 0.66 & 0.53 & 0.52 \\
 & SI & 0.67 & 0.51 & 0.27 \\
 & average & 0.61 & 0.51 & 0.39 \\\hline
\end{tabular}
\end{table}

\begin{table}[!htbp]
    \centering
    \caption{Definition for each icd code}
    \label{tab:icd_definition}
    \begin{tabular}{c|l}\hline
        \textbf{ICD-10/ICD-9 codes} & \textbf{Definition} \\ \hline
        Z60.2 & Problems related to living alone\\
        Z60.4 & Social exclusion and rejection\\
        Z60.9 & Problem related to social environment, unspecified\\
        Z63.2 & Inadequate family support \\
        Z63.3 & Absence of family member \\ \hline
        V60.3 & Person living alone\\
        V62.4 & Social Exclusion or Rejection \\ \hline
    \end{tabular}
    
\end{table}

\begin{landscape}
\begin{longtable}{|l|l|l|l|l|l|l|}\\\hline
\textbf{Sentence} & \textbf{GSA} & \textbf{RBS} & \textbf{LLM} & \textbf{Comment} & \textbf{\begin{tabular}[c]{@{}l@{}}Appearance \\ in accuracy \\ evaluation\end{tabular}} & \textbf{\begin{tabular}[c]{@{}l@{}}Post-Hoc \\ Ruling\end{tabular}} \\ \hline

\endfirsthead
\multicolumn{7}{c}%
{\tablename\ \thetable\ -- \textit{Continued from previous page}} \\ \hline

\textbf{Sentence} & \textbf{GSA} & \textbf{RBS} & \textbf{LLM} & \textbf{Comment} & \textbf{\begin{tabular}[c]{@{}l@{}}Appearance \\ in accuracy \\ evaluation\end{tabular}} & \textbf{\begin{tabular}[c]{@{}l@{}}Post-Hoc \\ Ruling\end{tabular}} \\ \hline\endhead

\hline \multicolumn{7}{c}{\textit{Continued on next page}} \\
\endfoot
\endlastfoot

\begin{tabular}[c]{@{}l@{}}His rent is paid for by \\ Catholic Charities \\ (until December 2019)\end{tabular} & - & - & \begin{tabular}[c]{@{}l@{}}SI: \\ instrumental \\ support\end{tabular} & \begin{tabular}[c]{@{}l@{}}"rent is paid for by" is \\ not in the Lexicon for \\ instrumental support \\ but perhaps should be \\ because this is a clear \\ example of \\ instrumental support \\ as we define it\end{tabular} & \begin{tabular}[c]{@{}l@{}}RBS correct; \\ LLM false \\ positive\end{tabular} & \begin{tabular}[c]{@{}l@{}}LLM is more \\ correct than \\ gold-standard \\ annotation\end{tabular} \\ \hline

\begin{tabular}[c]{@{}l@{}}pt is not close with sisters \\ who are all significantly \\ older than pt.\end{tabular} & - & - & \begin{tabular}[c]{@{}l@{}}SI: \\ no social \\ network\end{tabular} & \begin{tabular}[c]{@{}l@{}}the conservatie \\ approach does not \\ assume that the \\ patient lacks \\ connections outside \\ the sisters whereas \\ the LLM does not \\ consider that wider \\ context\end{tabular} & \begin{tabular}[c]{@{}l@{}}RBS correct; \\ LLM false \\ positive\end{tabular} & no change \\ \hline

\begin{tabular}[c]{@{}l@{}}attending supervisor \\ assisted with appeal \\ which was successful \\ in obtaining authorization\end{tabular} & - & - & \begin{tabular}[c]{@{}l@{}}SS: \\ instrumental \\ support\end{tabular} & \begin{tabular}[c]{@{}l@{}}LLM recognized the \\ instrumental support \\ happening, but the \\ support is towards the \\ note writer, not the \\ patient\end{tabular} & \begin{tabular}[c]{@{}l@{}}RBS correct; \\ LLM false \\ positive\end{tabular} & no change \\ \hline

\begin{tabular}[c]{@{}l@{}}social hx: {[}…{]} her only \\ family member is her \\ mother who is located \\ in Russia{[}…{]}\end{tabular} & - & - & \begin{tabular}[c]{@{}l@{}}SI: \\ no social\\  network\end{tabular} & \begin{tabular}[c]{@{}l@{}}while the rule indeed \\ is that having no \\ family counts as no \\ social network, in \\ this very long \\ complicated sentence \\ that was trimmed, the \\ annotators missed \\ this mention and \\ the RBS was unable \\ to identify\end{tabular} & \begin{tabular}[c]{@{}l@{}}RBS correct; \\ LLM false \\ positive\end{tabular} & \begin{tabular}[c]{@{}l@{}}LLM is more \\ correct than \\ gold-standard \\ annotation\end{tabular} \\ \hline

\begin{tabular}[c]{@{}l@{}}she was crying and \\ feeling hopeless went to \\ spend the night with \\ her mother.\end{tabular} & - & - & \begin{tabular}[c]{@{}l@{}}SI:\\ loneliness\end{tabular} & \begin{tabular}[c]{@{}l@{}}The LLM read an \\ implied feeling\end{tabular} & \begin{tabular}[c]{@{}l@{}}RBS correct; \\ LLM false \\ positive\end{tabular} & \begin{tabular}[c]{@{}l@{}}open to \\ interpretation\end{tabular} \\ \hline

I feel lonely "" & \begin{tabular}[c]{@{}l@{}}SI: \\ loneliness\end{tabular} & \begin{tabular}[c]{@{}l@{}}SI: \\ loneliness\end{tabular} & - & \begin{tabular}[c]{@{}l@{}}The irregular \\ punctuation \\ misled the LLM\end{tabular} & \begin{tabular}[c]{@{}l@{}}RBS correct; \\ LLM false \\ negative\end{tabular} & no change \\ \hline

\begin{tabular}[c]{@{}l@{}}Pt's three close friends\\  have died\end{tabular} & \begin{tabular}[c]{@{}l@{}}SI:\\ no social \\ network\end{tabular} & \begin{tabular}[c]{@{}l@{}}SS: \\ social \\ network\end{tabular} & \begin{tabular}[c]{@{}l@{}}SI:\\ no social \\ network\end{tabular} & \begin{tabular}[c]{@{}l@{}}the RBS could not \\ detect the tense in \\ this unfortunate \\ example\end{tabular} & \begin{tabular}[c]{@{}l@{}}RBS false \\ negative; \\ LLM true\end{tabular} & no change \\ \hline

\begin{tabular}[c]{@{}l@{}}describes feeling of \\ low-self esteem, guilt, \\ and isolation in this \\ time period.\end{tabular} & \begin{tabular}[c]{@{}l@{}}SI:\\ general\end{tabular} & - & - & \begin{tabular}[c]{@{}l@{}}isolation' was too \\ ambiguous of a term \\ to include alone in the \\ lexicon; unsure why\\  the LLM could not \\ recognize\end{tabular} & \begin{tabular}[c]{@{}l@{}}RBS false \\ negative; \\ LLM false \\ negative\end{tabular} & no change \\ \hline

\begin{tabular}[c]{@{}l@{}}When asked about his \\ living sitaution and social\\  support, the patient was \\ unable to answer.\end{tabular} & - & \begin{tabular}[c]{@{}l@{}}SS:\\ general\end{tabular} & - & \begin{tabular}[c]{@{}l@{}}A shortcoming of the \\ lexicon to pick up on \\ this out of context \\ mention of the phrase\end{tabular} & \begin{tabular}[c]{@{}l@{}}RBS false \\ positive; \\ LLM correct\end{tabular} & no change \\ \hline

\begin{tabular}[c]{@{}l@{}}He is part of a local \\ senior's walking group \\ and often hosts game \\ nights at his home\end{tabular} & \begin{tabular}[c]{@{}l@{}}SS: \\ social\\ network\end{tabular} & - & \begin{tabular}[c]{@{}l@{}}SS: \\ social \\ network\end{tabular} & \begin{tabular}[c]{@{}l@{}}not in lexicon, but a \\ real example that the \\ LLM can detect\end{tabular} & \begin{tabular}[c]{@{}l@{}}RBS false \\ negative; \\ LLM correct\end{tabular} & no change \\ \hline

\begin{tabular}[c]{@{}l@{}}Pt is a 27 yo f domiciled \\ with roommates with no \\ significant pmhx {[}…{]} who \\ presented to the ED with \\ EMS after roommates \\ called 911\end{tabular} & - & - & \begin{tabular}[c]{@{}l@{}}SS: \\ social \\ network\end{tabular} & \begin{tabular}[c]{@{}l@{}}a rule was made to \\ exclude living with \\ roomates and also \\ to consider the \\ person who \\ activated EMS as a \\ relationship; however \\ here it seems the \\ roommates are \\ important\end{tabular} & \begin{tabular}[c]{@{}l@{}}RBS correct; \\ LLM false \\ positive\end{tabular} & \begin{tabular}[c]{@{}l@{}}LLM is more \\ correct than \\ gold-standard \\ annotation\end{tabular} \\ \hline
\caption{Erroneous examples from Rule-based NLP systems (RBS) and LLM-based NLP systems.\label{tab:examples_both}}

\end{longtable}
\end{landscape}

\end{document}